\documentclass{article}
\usepackage{ijcai17}
\usepackage{times}
\usepackage{helvet}
\usepackage{courier}
\usepackage{textcomp}
\usepackage{amsmath,amssymb}
\usepackage{multicol}
\usepackage{float}
\usepackage{graphicx}
\usepackage{array}
\usepackage{multirow}
\usepackage{placeins}
\usepackage{subcaption}
\usepackage{color,soul}
\usepackage{url}
\usepackage{hyperref}
\usepackage{adjustbox}
\usepackage{caption}
\title{A Dual-Stage Attention-Based Recurrent Neural Network \\ for Time Series Prediction}
  \author{Yao Qin$^1$\thanks{Most of this work was performed while the first author was an intern at NEC Labs America.}, Dongjin Song$^2$, Haifeng Chen$^2$, Wei Cheng$^2$, Guofei Jiang$^2$, Garrison W.  Cottrell$^1$ \\
  $^1$University of California, San Diego\\    $^2$NEC Laboratories America, Inc.\\
  \{yaq007, gary\}@eng.ucsd.edu, \{dsong, Haifeng, weicheng, gfj\}@nec-labs.com}

\begin{document}
 \maketitle
\begin{abstract}
The Nonlinear autoregressive exogenous (NARX) model, which predicts the current value of a time series based upon its previous values as well as the current and past values of multiple driving (exogenous) series, has been studied for decades. Despite the fact that various NARX models have been developed, few of them can capture the long-term temporal dependencies appropriately and select the relevant driving series to make predictions. In this paper, we propose a dual-stage attention-based recurrent neural network (DA-RNN) to address these two issues. In the first stage, we introduce an input attention mechanism to adaptively extract relevant driving series (\textit{a.k.a.}, input features) at each time step by referring to the previous encoder hidden state. In the second stage, we use a temporal attention mechanism to select relevant encoder hidden states across all time steps. With this dual-stage attention scheme,
our model can not only make predictions effectively, but can also be easily interpreted. Thorough empirical studies based upon the SML 2010 dataset and the NASDAQ 100 Stock dataset demonstrate that the DA-RNN can outperform state-of-the-art methods for time series prediction.
\end{abstract}
\section{Introduction}
Time series prediction algorithms have been widely applied in many areas, \textit{e.g.}, financial market prediction~\cite{wu2013dynamic}, weather forecasting~\cite{chakraborty2012fine}, and complex dynamical system analysis~\cite{liu2015regularized}. Although the well-known autoregressive moving average (ARMA) model~\cite{Whittle1951} and its variants~\cite{asteriou2011arima,Brockwell2009} have shown their effectiveness for various real world applications, they cannot model nonlinear relationships and do not differentiate among the exogenous (driving) input terms. To address this issue, various nonlinear autoregressive exogenous (NARX) models~\cite{lin1996learning,gao2005narmax,Diaconescu08,Cottrell-NARX} have been developed. Typically, given the previous values of the target series, \textit{i.e.} $(y_1,y_2,\cdots,y_{t-1})$ with $y_{t-1}\in\mathbb{R}$, as well as the current and past values of $n$ driving (exogenous) series, \textit{i.e.}, $(\textbf{x}_1, \textbf{x}_2, \cdots, \textbf{x}_t)$ with $\textbf{x}_t\in\mathbb{R}^{n}$, the NARX model aims to learn a nonlinear mapping to the current value of target series $y_t$, \textit{i.e.}, $\hat{y}_t=F(y_1,y_2,\cdots,y_{t-1},\textbf{x}_1, \textbf{x}_2, \cdots, \textbf{x}_t)$, where $F(\cdot)$ is the mapping function to learn.

Despite the fact that a substantial  effort has been made for time series prediction via kernel methods~\cite{Chen2008}, ensemble methods~\cite{Bouchachia2008}, and Gaussian processes~\cite{Frigola2013}, the drawback is that most of these approaches employ a predefined nonlinear form and may not be able to capture the true underlying nonlinear relationship appropriately. Recurrent neural networks (RNNs)~\cite{rumelhart1986learning,werbos1990backpropagation,elman1991distributed}, a type of deep neural network specially designed for sequence modeling, have received a great amount of attention due to their flexibility in capturing nonlinear relationships. In particular, RNNs have shown their success in NARX time series forecasting in recent years~\cite{gao2005narmax,Diaconescu08}. Traditional RNNs, however, suffer from the problem of vanishing gradients~\cite{bengio1994learning} and thus have difficulty capturing long-term dependencies. Recently, long short-term memory units (LSTM)~\cite{hochreiter1997long} and the gated recurrent unit (GRU)~\cite{cho2014learning} have overcome this limitation and achieved great success in various applications, \textit{e.g.}, neural machine translation~\cite{bahdanau2014neural}, speech recognition~\cite{graves2013speech}, and image processing~\cite{karpathy2015deep}. Therefore, it is natural to consider state-of-the-art RNN methods, \textit{e.g.}, encoder-decoder networks~\cite{cho2014learning,sutskever2014sequence} and attention based encoder-decoder networks~\cite{bahdanau2014neural}, for time series prediction.

\begin{figure*}[t]

  \centering
  \includegraphics[height=5.34cm]{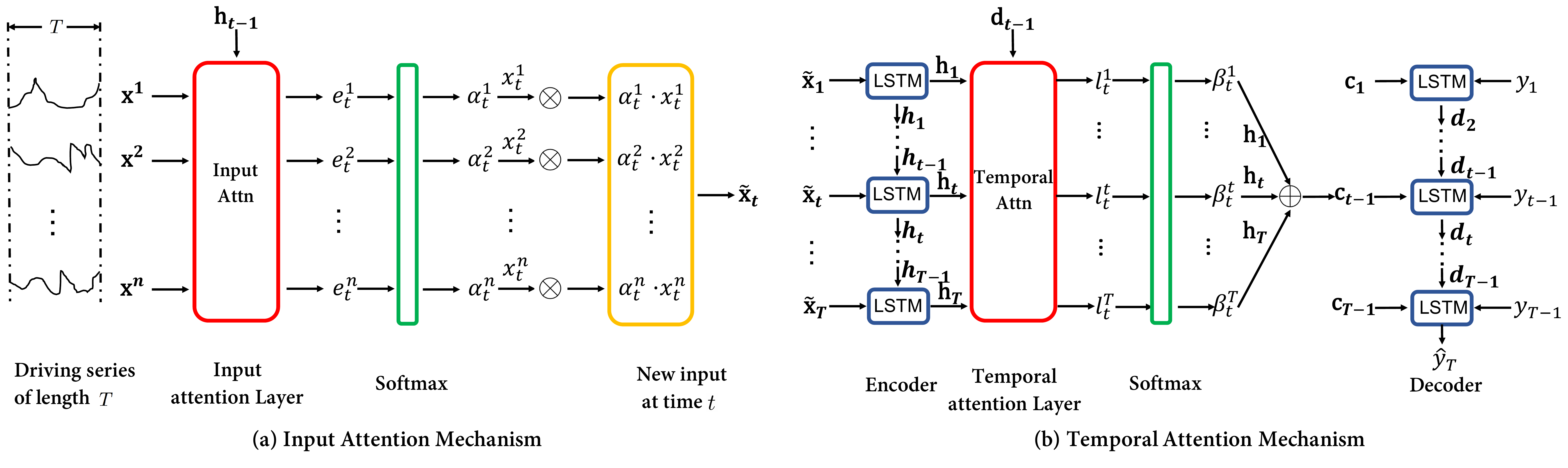}\\
  \vspace{-3mm}
  \caption{\small Graphical illustration of the dual-stage attention-based recurrent neural network. (a) The input attention mechanism computes the attention weights $\alpha^k_{t}$ for multiple driving series $\{\textbf{x}^1, \textbf{x}^2, \cdots, \textbf{x}^n\}$ conditioned on the previous hidden state $\textbf{h}_{t-1}$ in the encoder and then feeds the newly computed $\tilde{\textbf{x}}_t = (\alpha^1_tx^1_t, \alpha^2_tx^2_t, \cdots, \alpha^n_tx^n_t )^{\mathrm{\top}} $ into the encoder LSTM unit. (b) The temporal attention system computes  the attention weights $\beta^t_t$ based on the previous decoder hidden state $\textbf{d}_{t-1}$ and represents the input information as a weighted sum of the encoder hidden states across all the time steps. The generated context vector $\textbf{c}_{t}$ is then used as an input to the decoder LSTM unit. The output $\hat{y}_T$ of the last decoder LSTM unit is the predicted result.}\label{sm-show1}
\vspace{-5mm}
\end{figure*}

Based upon LSTM or GRU units, encoder-decoder networks ~\cite{kalchbrenner2013recurrent,cho2014properties,cho2014learning,sutskever2014sequence} have become popular due to their success in machine translation. The key idea is to encode the source sentence as a fixed-length vector and use the decoder to generate a translation. One problem with encoder-decoder networks is that their performance will deteriorate rapidly as the length of input sequence increases~\cite{cho2014properties}. In time series analysis, this could be a concern since we usually expect to make predictions based upon a relatively long segment of the target series as well as driving series. To resolve this issue, the attention-based encoder-decoder network~\cite{bahdanau2014neural} employs an attention mechanism to select parts of hidden states across all the time steps. Recently, a hierarchical attention network \cite{Yang2016}, which uses two layers of attention mechanism to select relevant encoder hidden states across all the time steps, was also developed. Although attention-based encoder-decoder networks and hierarchical attention networks have shown their efficacy for machine translation, image captioning~\cite{xu2015show}, and document classification, they may not be suitable for time series prediction. This is because when multiple driving (exogenous) series are available, the network cannot explicitly select relevant driving series
to make predictions. In addition, they have mainly been used for classification, rather than time series prediction.

To address these aforementioned issues, and inspired by some theories of human attention~\cite{hubner2010dual} that posit that human behavior is well-modeled by a two-stage attention mechanism, we propose a novel dual-stage attention-based recurrent neural network (DA-RNN) to perform time series prediction. In the first stage, we develop a new attention mechanism to adaptively extract the relevant driving series at each time step by referring to the previous encoder hidden state. In the second stage, a temporal attention mechanism is used to select relevant encoder hidden states across all time steps. These two attention models are well integrated within an LSTM-based recurrent neural network (RNN) and can be jointly trained using standard back propagation. In this way, the DA-RNN can adaptively select the most relevant input features as well as capture the long-term temporal dependencies of a time series appropriately. To justify the effectiveness of the DA-RNN, we compare it with state-of-the-art approaches using the SML 2010 dataset and the NASDAQ 100 Stock dataset with a large number of driving series. Extensive experiments not only demonstrate the effectiveness of the proposed approach, but also show that the DA-RNN is easy to interpret, and robust to noisy inputs.

\vspace{-3mm}
\section{Dual-Stage Attention-Based RNN}

In this section, we first introduce the notation we use in this work and the problem we aim to study. Then, we present the motivation and details of the DA-RNN for time series prediction.
\vspace{-3mm}
\subsection{Notation and Problem Statement}

Given $n$ driving series, \textit{i.e.}, $\textbf{X} = (\textbf{x}^1, \textbf{x}^2, \cdots, \textbf{x}^n)^{\top}= (\textbf{x}_1, \textbf{x}_2, \cdots, \textbf{x}_T)\in \mathbb{R}^{n\times T}$,  where $T$ is the length of window size, we use $\textbf{x}^{k}=(x^k_1,x^k_2,\cdot,x^k_T)^{\top}\in\mathbb{R}^T$ to represent a driving series of length $T$ and employ $\textbf{x}_t = (x^1_t, x^2_t, \cdots, x^n_t)^{\top} \in\mathbb{R}^n$ to denote a vector of $n$ exogenous (driving) input series at time $t$.

Typically, given the previous values of the target series, \textit{i.e.} $(y_1,y_2,\cdots,y_{T-1})$ with $y_t\in\mathbb{R}$, as well as the current and past values of $n$ driving (exogenous) series, \textit{i.e.}, $(\textbf{x}_1, \textbf{x}_2, \cdots, \textbf{x}_T)$ with $\textbf{x}_t\in\mathbb{R}^{n}$, the NARX model aims to learn a nonlinear mapping to the current value of the target series $y_T$:
\vspace{-3mm}
\begin{equation}
\begin{aligned}
\hat{y}_{T} = F(y_1, \cdots, y_{T-1}, \textbf{x}_1,  \cdots, \textbf{x}_T).
\end{aligned}
\end{equation}
where $F(\cdot)$ is a nonlinear mapping function we aim to learn.
\vspace{-2mm}
\subsection{Model}
Some theories of human attention~\cite{hubner2010dual} argue that behavioral results are best modeled by a two-stage attention mechanism. The first stage selects the elementary stimulus features while the second stage uses categorical information to decode the stimulus. Inspired by these theories, we propose a novel dual-stage attention-based recurrent neural network (DA-RNN) for time series prediction.
In the encoder, we introduce a novel input attention mechanism that can adaptively select the relevant driving series. In the decoder, a temporal attention mechanism is used to automatically select relevant encoder hidden states across all time steps. For the objective, a square loss is used. With these two attention mechanisms, the DA-RNN can adaptively select the most relevant input features and capture the long-term temporal dependencies of a time series. A  graphical illustration of the proposed model is shown in Figure~\ref{sm-show1}.

\subsubsection{Encoder with input attention}
The encoder is essentially an RNN that encodes the input sequences into a feature representation in machine translation~\cite{cho2014learning,sutskever2014sequence}. For time series prediction, given the input sequence $\textbf{X} = (\textbf{x}_1, \textbf{x}_2, \cdots, \textbf{x}_T)$ with $\textbf{x}_t\in\mathbb{R}^{n}$, where $n$ is the number of driving (exogenous) series, the encoder can be applied to learn a mapping from
$\textbf{x}_t$ to $\textbf{h}_t$ (at time step $t$) with
\vspace{0mm}
\begin{equation}\label{h1}
\textbf{h}_t = f_1(\textbf{h}_{t-1}, \textbf{x}_t),
\vspace{0mm}
\end{equation}
where $\textbf{h}_t\in \mathbb{R}^m$ is the hidden state of the encoder at time $t$, $m$ is the size of the hidden state, and $f_1$ is a non-linear activation function that could be an LSTM~\cite{hochreiter1997long} or gated recurrent unit (GRU)~\cite{cho2014learning}. In this paper, we use an LSTM unit as $f_1$ to capture long-term dependencies. Each LSTM unit has a memory cell with the state $\textbf{s}_t$ at time $t$. Access to the memory cell will be controlled by three sigmoid gates: forget gate $\textbf{f}_t$, input gate $\textbf{i}_t$ and output gate $\textbf{o}_t$. The update of an LSTM unit can be summarized as follows:
\begin{equation}
\textbf{f}_t = \sigma(\textbf{W}_f[\textbf{h}_{t-1};\textbf{x}_t] + \textbf{b}_f)
\end{equation}
\begin{equation}
\textbf{i}_t = \sigma(\textbf{W}_i[\textbf{h}_{t-1};\textbf{x}_t] + \textbf{b}_i)
\end{equation}
\begin{equation}
\textbf{o}_t = \sigma(\textbf{W}_o[\textbf{h}_{t-1};\textbf{x}_t] + \textbf{b}_o)
\end{equation}
\begin{equation}
\textbf{s}_t = \textbf{f}_t\odot\textbf{s}_{t-1} + \textbf{i}_t \odot \tanh(\textbf{W}_s[\textbf{h}_{t-1}; \textbf{x}_t]+\textbf{b}_s)
\end{equation}
\begin{equation}
\textbf{h}_t = \textbf{o}_t\odot\tanh(\textbf{s}_t)
\end{equation}
where $[\textbf{h}_{t-1};\textbf{x}_t] \in \mathbb{R}^{m+n}$ is a concatenation of the previous hidden state $\textbf{h}_{t-1}$ and the current input $\textbf{x}_t$. $\textbf{W}_f$, $\textbf{W}_i$, $\textbf{W}_o$, $\textbf{W}_s \in \mathbb{R}^{m\times (m+n)}$, and $\textbf{b}_f$, $\textbf{b}_i$, $\textbf{b}_o$, $\textbf{b}_s \in \mathbb{R}^{m}$ are parameters to learn. $\sigma$ and $\odot$ are a logistic sigmoid function and an element-wise multiplication, respectively. The key reason for using an LSTM unit is that the cell state sums activities over time, which can overcome the problem of vanishing gradients and better capture long-term dependencies of time series.

Inspired by the theory that the human attention system can select elementary stimulus features in the early stages of processing~\cite{hubner2010dual}, we propose an input attention-based encoder that can adaptively select the relevant driving series, which is of practical meaning in time series prediction.

Given the $k$-th input driving (exogenous) series $\textbf{x}^{k}=(x^k_1,x^k_2,\cdots,x^k_T)^{\top}\in\mathbb{R}^T$, we can construct an input attention mechanism via a deterministic attention model, \textit{i.e.}, a multilayer perceptron, by referring to the previous hidden state $\textbf{h}_{t-1}$ and the cell state $\textbf{s}_{t-1}$ in the encoder LSTM unit with:
\begin{equation}\label{input}
e_t^k = \textbf{v}^\top_e \tanh(\textbf{W}_e[\textbf{h}_{t-1};\textbf{s}_{t-1}] + \textbf{U}_e\textbf{x}^k)
\end{equation}
and
\vspace{-2mm}
\begin{equation}
\alpha^k_t = \frac{\exp(e^k_t)}{\sum^n_{i=1}\exp(e^i_t)},
\end{equation}
where $\textbf{v}_e \in\mathbb{R}^T$, $\textbf{W}_e \in\mathbb{R}^{T\times 2m}$ and $\textbf{U}_e\in\mathbb{R}^{T\times T}$ are parameters to learn.We omit the bias terms in Eqn.~(\ref{input}) to be succinct. $\alpha_t^k$ is the attention weight measuring the importance of the $k$-th input feature (driving series) at time $t$. A softmax function is applied to $e_t^k$ to ensure all the attention weights sum to 1. The input attention mechanism is a feed forward network that can be jointly trained with other components of the RNN. With these attention weights, we can adaptively extract the driving series with
\vspace{-2mm}
\begin{equation}
\tilde{\textbf{x}}_t =  (\alpha^1_tx^1_t, \alpha^2_tx^2_t, \cdots, \alpha^n_tx^n_t )^{\mathrm{\top}}.
\end{equation}
Then the hidden state at time $t$ can be updated as:
\begin{equation}
\textbf{h}_t = f_1(\textbf{h}_{t-1},\tilde{\textbf{x}}_t),
\end{equation}
where $f_1$ is an LSTM unit that can be computed according to Eqn.~(3) - (7) with $\textbf{x}_t$ replaced by the newly computed $\tilde{\textbf{x}}_t$. With the proposed input attention mechanism, the encoder can selectively focus on certain driving series rather than treating all the input driving series equally.

\subsubsection{Decoder with temporal attention}
To predict the output $\hat{y}_{T}$, we use another LSTM-based recurrent neural network to decode the encoded input information. However, as suggested by \citeauthor{cho2014properties}~[\citeyear{cho2014properties}], the performance of the encoder-decoder network can deteriorate rapidly as the length of the input sequence increases. Therefore, following the encoder with input attention, a temporal attention mechanism is used in the decoder to adaptively select relevant encoder hidden states across all time steps. Specifically, the attention weight of each encoder hidden state at time $t$ is calculated based upon the previous decoder hidden state $\textbf{d}_{t-1}\in\mathbb{R}^p$ and the cell state of the LSTM unit $\textbf{s}'_{t-1} \in \mathbb{R}^p$  with
\begin{equation}
l_t^i = \textbf{v}_d^\top \tanh(\textbf{W}_d[\textbf{d}_{t-1};\textbf{s}'_{t-1}] +
\textbf{U}_d\textbf{h}_i), \quad 1\leq i \leq T
\end{equation}
and
\vspace{-2mm}
\begin{equation}
\beta_t^i = \frac{\exp(l_t^i)}{\sum^T_{j=1}\exp(l_t^j)},
\end{equation}
where $[\textbf{d}_{t-1};\textbf{s}'_{t-1}]\in\mathbb{R}^{2p}$ is a concatenation of the previous hidden state and cell state of the LSTM unit. $\textbf{v}_d \in\mathbb{R}^m$, $\textbf{W}_d \in\mathbb{R}^{m\times 2p}$ and $\textbf{U}_d\in\mathbb{R}^{m\times m}$ are parameters to learn. The bias terms here have been omitted for clarity. The attention weight $\beta_t^i$ represents the importance of the $i$-th encoder hidden state for the prediction. Since each encoder hidden state $\textbf{h}_i$ is mapped to a temporal component of the input, the attention mechanism computes the context vector $\textbf{c}_t$ as a weighted sum of all the encoder hidden states $\{\textbf{h}_1, \textbf{h}_2, \cdots, \textbf{h}_T\}$,
\vspace{-2mm}
\begin{equation}
\textbf{c}_t = \sum^T_{i=1}\beta^i_{t}\textbf{h}_i.
\end{equation}
Note that the context vector $\textbf{c}_t$ is distinct at each time step.

Once we get the weighted summed context vectors, we can combine them with the given target series $(y_1,y_2,\cdots,y_{T-1})$:
\vspace{-2mm}
\begin{equation}
{\tilde{y}}_{t-1} = \tilde{\textbf{w}}^\top[y_{t-1}; \textbf{c}_{t-1}] + \tilde{b},
\end{equation}
where $[y_{t-1}; \textbf{c}_{t-1}]\in\mathbb{R}^{m+1}$ is a concatenation of the decoder input $y_{t-1}$ and the computed context vector $\textbf{c}_{t-1}$. Parameters $\tilde{\textbf{w}}\in\mathbb{R}^{m+1}$ and $\tilde{b}\in\mathbb{R}$ map the concatenation to the size the decoder input. The newly computed
${\tilde{y}}_{t-1}$ can be used for the update of the decoder hidden state at time $t$:
\begin{equation}
\textbf{d}_t = f_2(\textbf{d}_{t-1}, \tilde{y}_{t-1}).
\end{equation}
We choose the nonlinear function $f_2$ as an LSTM unit~\cite{hochreiter1997long}, which has been widely used in modeling long-term dependencies. Then $\textbf{d}_t$ can be updated as:
\begin{equation}
\textbf{f}'_t = \sigma(\textbf{W}'_f[\textbf{d}_{t-1};{\tilde{y}}_{t-1}] + \textbf{b}'_f)
\end{equation}
\begin{equation}
\textbf{i}'_t = \sigma(\textbf{W}'_i[\textbf{d}_{t-1};{\tilde{y}}_{t-1}] + \textbf{b}'_i)
\end{equation}
\begin{equation}
\textbf{o}'_t = \sigma(\textbf{W}'_o[\textbf{d}_{t-1};{\tilde{y}}_{t-1}] + \textbf{b}'_o)
\end{equation}
\begin{equation}
\textbf{s}'_t = \textbf{f}'_t\odot\textbf{s}'_{t-1} + \textbf{i}'_t \odot \tanh(\textbf{W}'_s[\textbf{d}_{t-1};{\tilde{y}}_{t-1}]+\textbf{b}'_s)
\end{equation}
\begin{equation}
\textbf{d}_t = \textbf{o}'_t\odot\tanh(\textbf{s}'_t)
\end{equation}
where $[\textbf{d}_{t-1};{\tilde{y}}_{t-1}]\in\mathbb{R}^{p+1}$ is a concatenation of the previous hidden state $\textbf{d}_{t-1}$ and the decoder input $\tilde{y}_{t-1}$. $\textbf{W}'_f$, $\textbf{W}'_i$, $\textbf{W}'_o$, $\textbf{W}'_s \in \mathbb{R}^{p\times (p+1)}$, and $\textbf{b}'_f$, $\textbf{b}'_i$, $\textbf{b}'_o$, $\textbf{b}'_s \in \mathbb{R}^{p}$ are parameters to learn. $\sigma$ and $\odot$ are a logistic sigmoid function and an element-wise multiplication, respectively.

For NARX modeling, we aim to use the DA-RNN to approximate the function $F$ so as to obtain an estimate of the current output $\hat{y}_{T}$ with the observation of all inputs as well as previous outputs. Specifically, $\hat{y}_{T}$ can be obtained with
\begin{equation} \label{eq1}
\begin{split}
\hat{y}_{T} & = F(y_1, \cdots, y_{T-1}, \textbf{x}_1,  \cdots, \textbf{x}_T)\\
 &  = \textbf{v}_y^\top (\textbf{W}_y[\textbf{d}_T; \textbf{c}_{T}] + \textbf{b}_w) + b_v,\\
\end{split}
\end{equation}
where $[\textbf{d}_T; \textbf{c}_T] \in \mathbb{R}^{p+m}$ is a concatenation of the decoder hidden state and the context vector. The parameters $\textbf{W}_y \in \mathbb{R} ^{p \times {(p+m)}}$ and $\textbf{b}_w \in\mathbb{R}^p$ map the concatenation to the size of the decoder hidden states.  The linear function with weights $\textbf{v}_y\in\mathbb{R}^p$ and bias $b_v\in\mathbb{R}$ produces the final prediction result.
\vspace{-2mm}

\begin{table}[t]
\centering
\caption{\small The statistics of two datasets.}\vspace{-2mm}
\label{datasets}
\begin{tabular}{|c|c|c|c|c|c|}
\hline\hline
\multicolumn{1}{|c|}{\multirow{2}{*}{Dataset}} & \multirow{2}{*}{\begin{tabular}[c]{@{}c@{}}driving \\ series\end{tabular}} & \multicolumn{4}{c|}{size}                     \\ \cline{3-6}
\multicolumn{1}{|c|}{}                              &                                                                                      & train   & valid   & \multicolumn{2}{l|}{test} \\ \hline\hline
SML 2010                                       & 16                                                                                   & 3,200   & 400    & \multicolumn{2}{l|}{537} \\ \hline
NASDAQ 100 Stock                                          & 81                                                                                   & 35,100   & 2,730     & \multicolumn{2}{l|}{2,730}  \\ \hline\hline
\end{tabular}
\end{table}

\vspace{-2mm}
\subsubsection{Training procedure}
We use minibatch stochastic gradient descent (SGD) together with the Adam optimizer~\cite{kingma2014adam} to train the model. The size of the minibatch is 128. The learning rate starts from 0.001 and is reduced by 10\% after each 10000 iterations.
The proposed DA-RNN is smooth and differentiable, so the parameters can be learned by standard back propagation with mean squared error as the objective function:
\vspace{-2mm}
\begin{equation}
\mathcal{O}(y_T,\hat{y}_{T}) = \frac{1}{N}\sum^N_{i=1}(\hat{y}_T^i - y_T^i)^2,
\vspace{-2mm}
\end{equation}
where $N$ is the number of training samples. We implemented the DA-RNN in the Tensorflow framework~\cite{abadi2015tensorflow}.
\vspace{-3mm}
\section{Experiments}
In this section, we first describe two datasets for empirical studies. Then, we introduce the parameter settings of DA-RNN and the evaluation metrics. Finally, we compare the proposed DA-RNN against four different baseline methods, interpret the input attention as well as the temporal attention of DA-RNN, and study its parameter sensitivity.

\vspace{-2mm}
\subsection{Datasets and Setup}
To test the performance of different methods for time series prediction, we use two different datasets as shown in Table \ref{datasets}.

SML 2010 is a public dataset used for indoor temperature forecasting.
This dataset is collected from a monitor system mounted in a domestic house. We use the room temperature as the target series and select 16 relevant driving series that contain approximately 40 days of monitoring data. The data was sampled every minute and was smoothed with 15 minute means. In our experiment, we use the first $3200$ data points as the training set, the following $400$ data points as the validation set, and the last $537$ data points as the test set.

In the NASDAQ 100 Stock dataset\footnote{http://cseweb.ucsd.edu/$\sim$yaq007/NASDAQ100$\_$stock$\_$data.html}, we collected the stock prices of 81 major corporations under NASDAQ 100, which are used as the driving time series. The index value of the NASDAQ 100 is used as the target series. The frequency of the data collection is minute-by-minute. This data covers the period from July 26, 2016 to December 22, 2016,  105 days in total. Each day contains 390 data points from the opening to closing of the market except that there are 210 data points on November 25 and 180 data points on December 22. In our experiments, we use the first 35,100 data points as the training set and the following 2,730 data points as the validation set. The last 2,730 data points are used as the test set. This dataset is publicly available and will be continuously enlarged to aid the research in this direction.

\begin{figure}[t]\vspace{0mm}
\centering
    \includegraphics[width=0.48\textwidth]{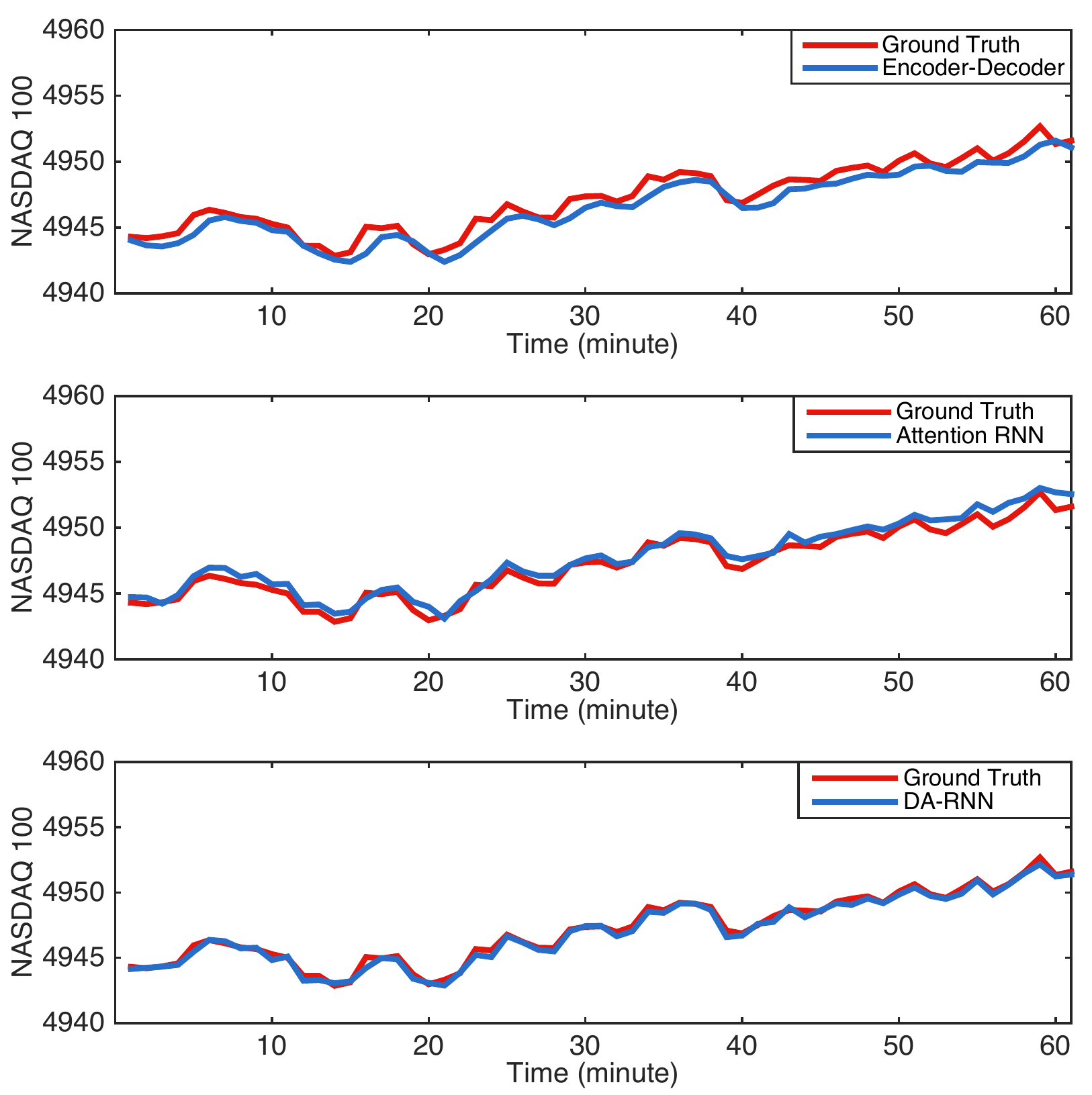}\\

    \vspace{-1mm}
    \caption{\small NASDAQ 100 Index vs. Time. Encoder-Decoder (top) and Attention RNN (middle), are compared with DA-RNN (bottom). }
    \label{fig:2}
  \vspace{-3mm}
\end{figure}

\begin{table*}[t]
\centering
\caption{\small Time series prediction results over the SML 2010 Dataset and NASDAQ 100 Stock Dataset (best performance displayed in \textbf{boldface}). The size of encoder hidden states $m$ and decoder hidden states $p$ are set as $m=p=64$ and $128$.}\vspace{-1mm}
\label{table:3}
\vspace{-2mm}
\begin{adjustbox}{max width=0.99\textwidth}
\begin{tabular}{|c|c|c|c|c|c|c|}
\hline\hline

\multirow{4}{*}{Models} & \multicolumn{3}{c|}{\multirow{2}{*}{\textbf{SML 2010 Dataset}} }                                                                                                                                                                            & \multicolumn{3}{c|}{\multirow{2}{*}{\textbf{NASDAQ 100 Stock Dataset}} }                                                                      \\
    & \multicolumn{3}{c|}{}                                                                                                                                                                                                                                                             & \multicolumn{3}{l|}{}      \\\cline{2-7}
                        & \multirow{2}{*}{\begin{tabular}[c]{@{}c@{}}$\textbf{MAE}$\\ ($\times 10^{-2}\%$)\end{tabular}} & \multirow{2}{*}{\begin{tabular}[c]{@{}c@{}}$\textbf{MAPE}$\\ ($\times 10^{-2}\%$)\end{tabular}} & \multirow{2}{*}{\begin{tabular}[c]{@{}c@{}}$\textbf{RMSE}$\\ ($\times 10^{-2}\%$)\end{tabular}} & \multirow{2}{*}{$\textbf{MAE}$} & \multirow{2}{*}{\begin{tabular}[c]{@{}c@{}}$\textbf{MAPE}$\\ ($\times 10^{-2}\%$)\end{tabular}} & \multirow{2}{*}{$\textbf{RMSE}$} \\
                         &                                                                     &                                                                      &                                                                      &                      &                                                                      &                       \\ \hline\hline
ARIMA [\citeyear{asteriou2011arima}]                                  & 1.95                 & 9.29                 & 2.65 & 0.91             & 1.84             & 1.45          \\ \hline
NARX RNN [\citeyear{Diaconescu08}]    & 1.79$\pm 0.07$                 & 8.64$\pm0.29$                 &2.34$\pm 0.08$                                                                        & 0.75$\pm 0.09$              & 1.51 $\pm 0.17$                & 0.98$\pm 0.10$                \\ \hline
Encoder-Decoder (64) [\citeyear{cho2014learning}]    & 2.59$ \pm  0.07$                & 12.1$ \pm 0.34$                    & 3.37$ \pm  0.07$                                                     &0.97$ \pm  0.06$                   & 1.96$ \pm  0.12$                     & 1.27$ \pm  0.05$                   \\ \hline
Encoder-Decoder (128) [\citeyear{cho2014learning}]     & 1.91$\pm 0.02$                 & 9.00$\pm 0.10$                 & 2.52$\pm 0.04$                                                 & 0.72 $\pm 0.03$              &1.46$\pm 0.06$                & 1.00$\pm 0.03$                    \\ \hline
Attention RNN (64) [\citeyear{bahdanau2014neural}]    & 1.78$\pm 0.03$                 & 8.46$\pm 0.09$                 & 2.32$\pm 0.03$                                          & 0.76$\pm 0.08$              & 1.54$\pm 0.02$   & 1.00$\pm 0.09$                  \\ \hline
Attention RNN (128) [\citeyear{bahdanau2014neural}]  & 1.77$\pm 0.02$                  & 8.45$\pm 0.09$                  & 2.33$\pm 0.03$                                           & 0.71$\pm 0.05$                & 1.43$\pm 0.09$                & 0.96$\pm 0.05$                  \\ \hline\hline
Input-Attn-RNN (64)          &   1.88$\pm 0.04$                 & 8.89$\pm 0.19$                 & 2.50$\pm 0.05$                                                        & 0.28$\pm 0.02$                  & 0.57$\pm 0.04$                 & 0.41$\pm 0.03$              \\ \hline
Input-Attn-RNN (128)       & 1.70$\pm 0.03$                 & 8.09$\pm 0.15$   & 2.24$\pm 0.03$                           & 0.26$\pm 0.02$              & 0.53$\pm 0.03$              & 0.39$\pm 0.03$             \\ \hline\hline
DA-RNN (64)      & 1.53$\pm 0.01$                  & 7.31$\pm 0.05$            & 2.02$\pm 0.01$                       & \textbf{{0.21$\pm$ 0.002} }          & \textbf{{0.43$\pm$ 0.005} }          & \textbf{{0.31$\pm$ 0.003 } }              \\ \hline
DA-RNN (128)                & \multicolumn{1}{c|}{\textbf{{1.50$\pm$ \textbf{0.01}}}} & \multicolumn{1}{c|}{\textbf{{7.14$\pm$ 0.07}}} & \multicolumn{1}{c|}{\textbf{{1.97$\pm$ 0.01}}}           & \multicolumn{1}{c|}{0.22$\pm$ 0.002} & \multicolumn{1}{c|}{0.45$\pm$ 0.005} & \multicolumn{1}{c|}{0.33$\pm$ 0.003}   \\ \hline
 \hline

\end{tabular}
\end{adjustbox}
\vspace{-3mm}
\end{table*}
\vspace{-2mm}
\subsection{Parameter Settings and Evaluation Metrics}
There are three parameters in the DA-RNN, \textit{i.e.}, the number of time steps in the window $T$, the size of hidden states for the encoder $m$, and the size of hidden states for the decoder $p$. To determine the window size $T$, we conducted a grid search over $T\in\{3, 5, 10, 15, 25\}$. The one ($T=10$) that achieves the best performance over validation set is used for test. For the size of hidden states for encoder ($m$) and decoder ($p$), we set $m=p$ for simplicity and conduct grid search over $m=p\in\{16,32,64,128,256\}$. Those two (\textit{i.e}, $m=p=64$, $128$) that achieve the best performance over the validation set are used for evaluation. For all the RNN based approaches (\textit{i.e.}, NARX RNN, Encoder-Decoder, Attention RNN, Input-Attn-RNN and DA-RNN), we train them 10 times and report their average performance and standard deviations for comparison.

To measure the effectiveness of various methods for time series prediction, we consider three different evaluation metrics. Among them, root mean squared error (RMSE)~\cite{plutowski1996experience} and mean absolute error (MAE) are two scale-dependent measures, and mean absolute percentage error (MAPE) is a scale-dependent measure. Specifically, assuming $y_t$ is the target at time $t$ and $\hat{y}_t$ is the predicted value at time $t$, RMSE is defined as $\textbf{RMSE} = \sqrt{\frac{1}{N}\sum^N_{i=1}(y_t^i - \hat{y}_t^i)^2}$ and MAE is denoted as $\textbf{MAE} = \frac{1}{N}\sum^N_{i=1}|y_t^i - \hat{y}_t^i|$. When comparing the prediction performance across different datasets, mean absolute percentage error is popular because it measures the prediction deviation proportion in terms of the true values, \textit{i.e.},
$\textbf{MAPE} = \frac{1}{N}\sum^N_{i=1}|\frac{y^i_t - \hat{y}^i_t}{y^i_t}|\times 100\%$.
\vspace{-1mm}
\subsection{Results-I: Time Series Prediction}
To demonstrate the effectiveness of the DA-RNN, we compare it against 4 baseline methods. Among them, the autoregressive integrated moving average (ARIMA) model is a generalization of an autoregressive moving average (ARMA) model~\cite{asteriou2011arima}. NARX recurrent neural network (NARX RNN) is a classic method to address time series prediction~\cite{Diaconescu08}. The encoder-decoder network (Encoder-Decoder)~\cite{cho2014learning} and attention-based encoder-decoder network (Attention RNN)~\cite{bahdanau2014neural} were originally used for machine translation tasks, in which each time step of the decoder output should be used to produce a probability distribution over the translated word codebook. To perform time series prediction, we modify these two approaches by changing the output to be a single scalar value, and use a squared loss as the objective function (as we did for the DA-RNN). The input to these networks is no longer words or word representations, but the $n$ scalar driving series of length $T$. Additionally, the decoder has the additional input of the previous values of the target series as the given information.

Furthermore, we show the effectiveness of DA-RNN via step-by-step justification. Specifically, we compare dual-stage attention-based recurrent neural network (DA-RNN) against the setting that only employs its input attention mechanism (Input-Attn-RNN). For all RNN-based methods, the encoder takes $n$ driving series of length $T$ as the input and the decoder takes the previous values of the target series as the given information for fair comparison.
\begin{figure}[!tbp]

    \includegraphics[width=0.48\textwidth, height=0.22\textwidth]{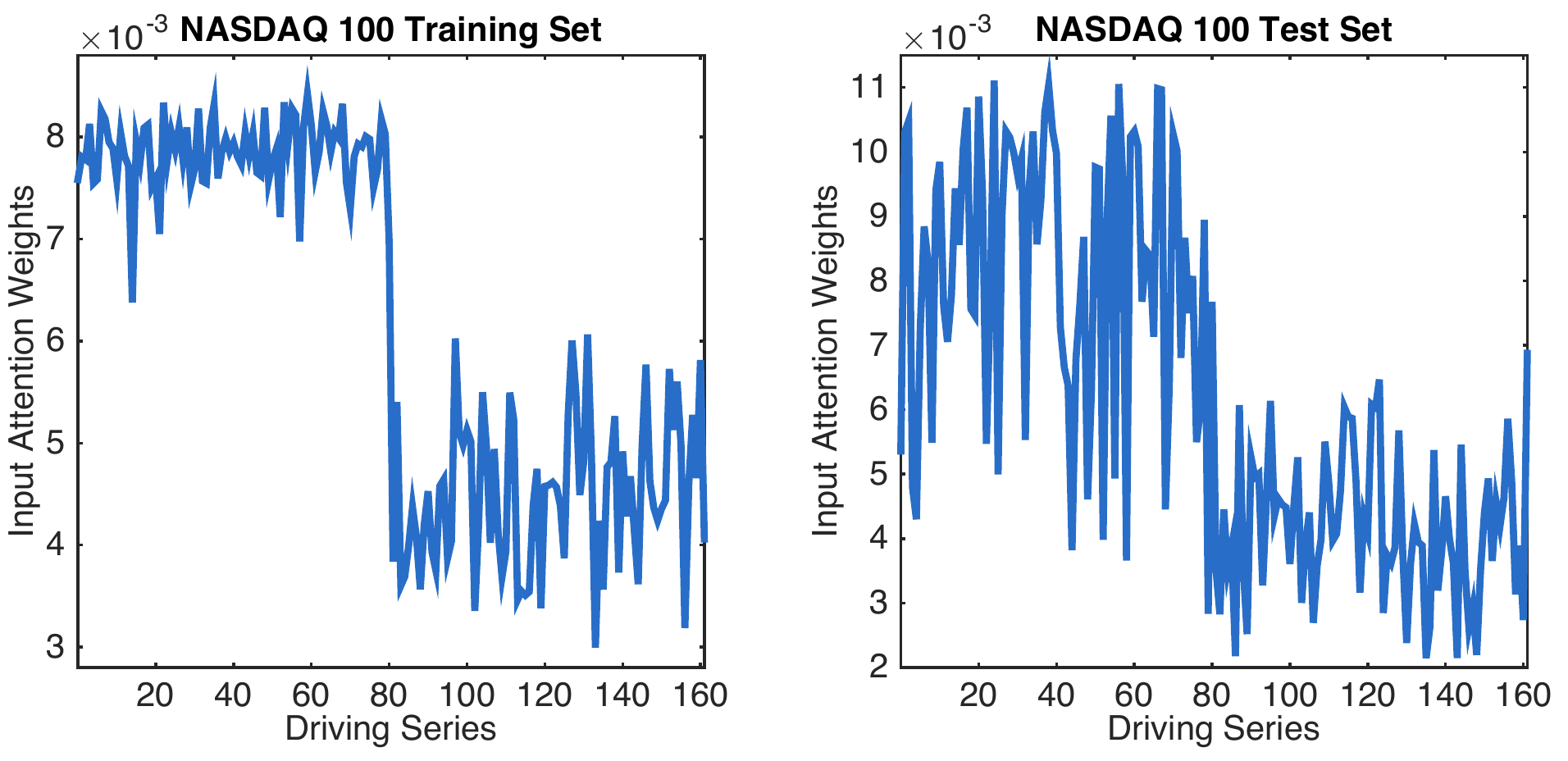}
    \vspace{-7mm}
  \caption{\small Plot of the input attention weights for DA-RNN from a single encoder time step. The first 81 weights are on 81 original driving series and the last 81 weights are on 81 noisy driving series. (left) Input attention weights on NASDAQ100 training set. (right) Input attention weights on NASDAQ100 test set. }\label{fig:3}
  \vspace{-5.5mm}
\end{figure}
The time series prediction results of DA-RNN and baseline methods over the two datasets are shown in Table~\ref{table:3}.

In Table \ref{table:3}, we observe that the \textbf{RMSE} of ARIMA is generally worse than RNN based approaches. This is because ARIMA only considers the target series $(y_1,\cdots,y_{t-1})$ and ignores the driving series $(\textbf{x}_{1},\cdots, \textbf{x}_{t})$. For RNN based approaches, the performance of NARX RNN and Encoder-Decoder are comparable. Attention RNN generally outperforms Encoder-Decoder since it is capable to select relevant hidden states across all the time steps in the encoder. Within DA-RNN, the input attention RNN (Input-Attn-RNN (128)) consistently outperforms Encoder-Decoder as well as Attention RNN. This suggests that adaptively extracting driving series can provide more reliable input features to make accurate predictions. With integration of the input attention mechanism as well as temporal attention mechanism, our DA-RNN achieves the best \textbf{MAE}, \textbf{MAPE}, and \textbf{RMSE} across two datasets since it not only uses an input attention mechanism to extract relevant driving series, but also employs a temporal attention mechanism to select relevant hidden features across all time steps.

For visual comparison, we show the prediction result of Encoder-Decoder ($m=p=128$), Attention RNN ($m=p=128$) and DA-RNN ($m=p=64$) over the NASDAQ 100 Stock dataset in Figure~\ref{fig:2}. We observe that DA-RNN generally fits the ground truth much better than Encoder-Decoder and Attention RNN.

\vspace{-3mm}
\subsection{Results-II: Interpretation}
\vspace{-1mm}
\begin{figure}[t]
  \centering
  \includegraphics[width=0.5\textwidth,height=0.21\textwidth]{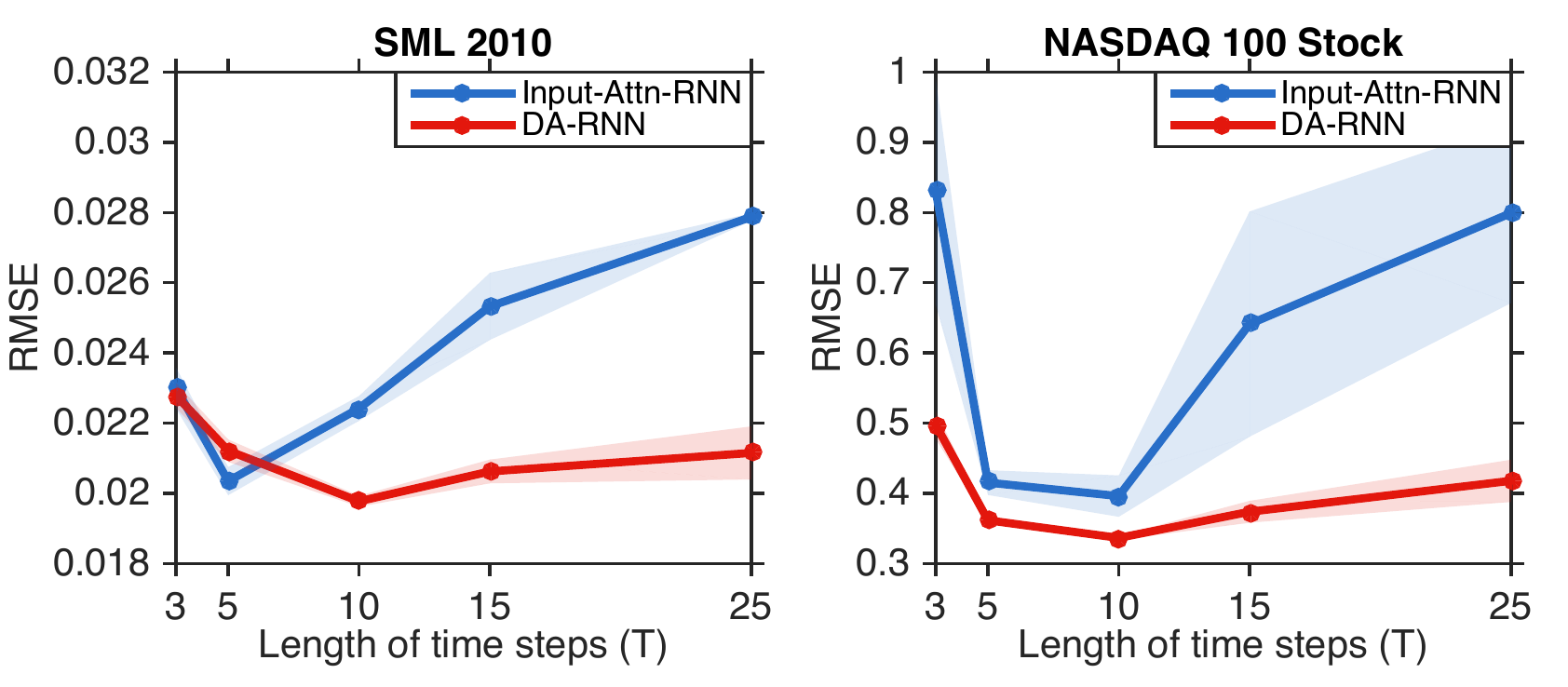}
  \vspace{-6.8mm}
  \caption{\small {\textbf{RMSE} vs. length of time steps $T$ over SML 2010 (left) and NASDAQ 100 Stock (right).}}\label{fig:4}
 \vspace{-5.5mm}
\end{figure}

To study the effectiveness of the input attention mechanism within DA-RNN, we test it with noisy driving (exogenous) series as the input. Specifically, within NASDAQ 100 Stock dataset, we generate 81 additional noisy driving series by randomly permuting the original 81 driving series. Then, we put these 81 noisy driving series together with the 81 original driving series as the input and test the effectiveness of DA-RNN. When the length of time steps $T$ is 10 and the size of hidden states is $m=p=128$, DA-RNN achieves \textbf{MAE}: $0.28 \pm 0.007$, \textbf{MAPE}: (0.56 $\pm$0.01)$\times 10^{-2}$ and \textbf{RMSE}: $0.42 \pm 0.009$, which are comparable to its performance in Table \ref{table:3}. This indicates that DA-RNN is robust to noisy inputs.

To further investigate the input attention mechanism, we plot the input attention weights of DA-RNN for the 162 input driving series (the first 81 are original and the last 81 are noisy) in Figure \ref{fig:3}. The plotted attention weights in Figure~\ref{fig:3} are taken from a single encoder time step and similar patterns can also be observed for other time steps. We find that the input attention mechanism can automatically assign larger weights for the 81 original driving series and smaller weights for the 81 noisy driving series in an online fashion using the activation of the input attention network to scale these weights. This demonstrates that input attention mechanism can aid DA-RNN to select relevant input driving series and suppress noisy input driving series.

To investigate the effectiveness of the temporal attention mechanism within DA-RNN, we compare DA-RNN to Input-Attn-RNN when the length of time steps $T$ varies from 3, 5, 10, 15, to 25. The detailed results over two datasets are shown in Figure 4. We observe that when $T$ is relatively large, DA-RNN can significantly outperform Input-Attn-RNN. This suggests that temporal attention mechanism can capture long-term dependencies by selecting relevant encoder hidden states across all the time steps.
\vspace{-2mm}
\subsection{Results-III: Parameter Sensitivity}
\vspace{0mm}
\begin{figure}[t]
  \centering
  \vspace{-0.5mm}
   \includegraphics[width=0.5\textwidth,height=0.21\textwidth]{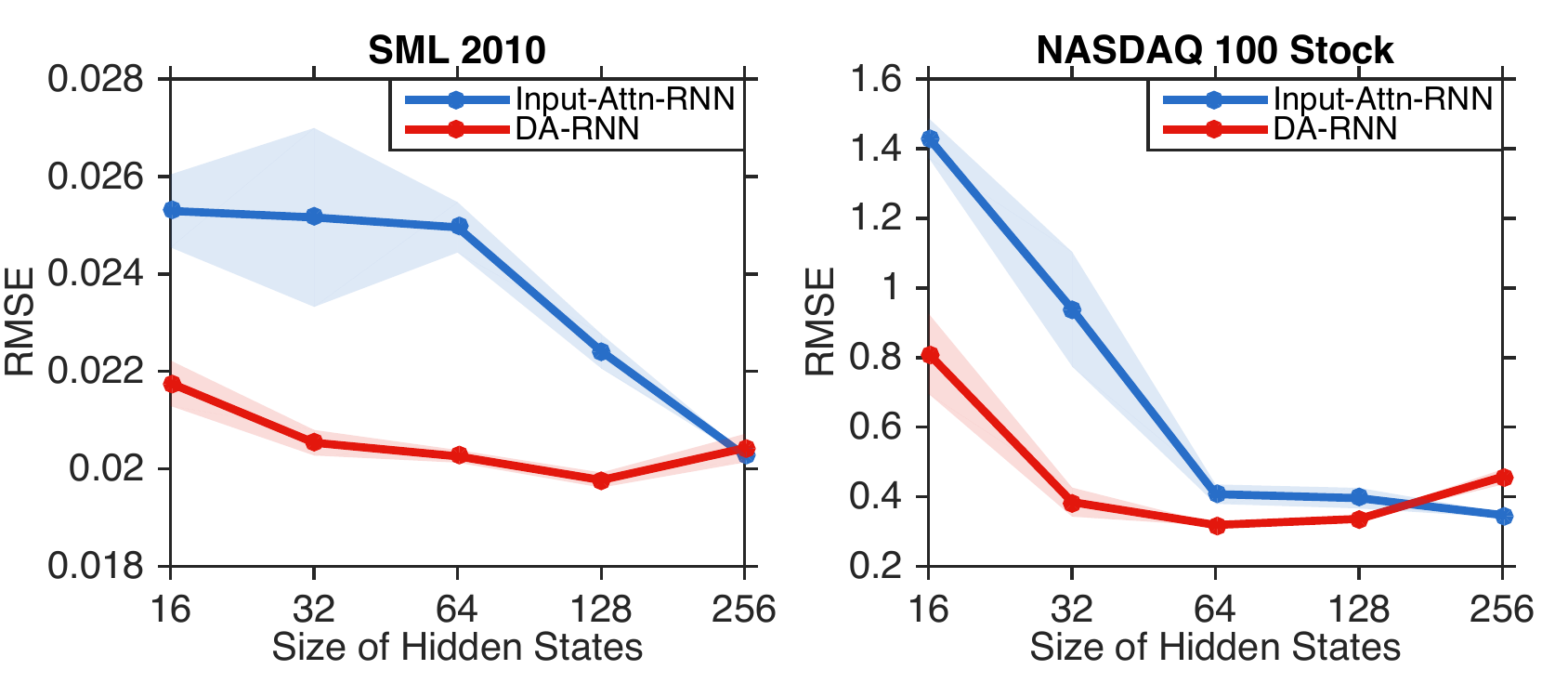}
  \vspace{-6.2mm}
  \caption{\small \textbf{RMSE} vs. size of hidden states of encoder/decoder over SML 2010 (left) and NASDAQ 100 Stock (right).}\label{fig:5}
  \vspace{-3.5mm}
\end{figure}

We study the sensitivity of DA-RNN with respect to its parameters, \textit{i.e.}, the length of time steps $T$ and the size of hidden states for encoder $m$ (decoder $p$). When we vary $T$ or $m$ ($p$), we keep the others fixed. By setting $m=p=128$, we plot the \textbf{RMSE} versus different lengths of time steps in the window $T$ in Figure 4. It is easily observed that the performance of DA-RNN and Input-Attn-RNN will be worse when the length of time steps is too short or too long while DA-RNN is relatively more robust than Input-Attn-RNN. By setting $T=10$, we also plot the \textbf{RMSE} versus different sizes of hidden states for encoder and decoder ($m=p\in\{16,32,64,128,256\}$) in Figure 5. We notice that DA-RNN usually achieves the best performance when $m=p=64$ or $128$. Moreover, we can also conclude that DA-RNN is more robust to parameters than Input-Attn-RNN.
\vspace{-2mm}
\section{Conclusion}
In this paper, we proposed a novel dual-stage attention-based recurrent neural network (DA-RNN), which consists of an encoder with an input attention mechanism and a decoder with a temporal attention mechanism. The newly introduced input attention mechanism can adaptively select the relevant driving series. The temporal attention mechanism can naturally capture the long-range temporal information of the encoded inputs. Based upon these two attention mechanisms, the DA-RNN can not only adaptively select the most relevant input features, but can also capture the long-term temporal dependencies of a time series appropriately. Extensive experiments on the SML 2010 dataset and the NASDAQ 100 Stock dataset demonstrated that our proposed DA-RNN can outperform state-of-the-art methods for time series prediction.

The proposed dual-stage attention-based recurrent neural network (DA-RNN) not only can be used for time series prediction, but also has the potential to serve as a general feature learning tool in computer vision tasks~\cite{pu2016adaptive,qin2015saliency}. In the future, we are going to employ DA-RNN to perform ranking and binary coding~\cite{song2015top,song2016fast}.

\vspace{-2mm}
\section*{Acknowledgments}
GWC is supported in part by NSF cooperative agreement SMA 1041755 to the Temporal Dynamics of Learning Center, and a gift from Hewlett Packard. GWC and YQ were also partially supported by Guangzhou Science and Technology Planning Project (Grant No. 201704030051).

\newpage
\bibliographystyle{named}

\small
\begin{thebibliography}{}

\bibitem[\protect\citeauthoryear{Abadi \bgroup \em et al.\egroup
  }{2015}]{abadi2015tensorflow}
Mart{\i}n Abadi, Ashish Agarwal, Paul Barham, Eugene Brevdo, Zhifeng Chen,
  Craig Citro, Greg~S Corrado, Andy Davis, Jeffrey Dean, Matthieu Devin, et~al.
\newblock Tensorflow: Large-scale machine learning on heterogeneous systems.
\newblock 2015.

\bibitem[\protect\citeauthoryear{Asteriou and Hall}{2011}]{asteriou2011arima}
Dimitros Asteriou and Stephen~G Hall.
\newblock Arima models and the box-jenkins methodology.
\newblock {\em Applied Econometrics}, 2(2):265--286, 2011.

\bibitem[\protect\citeauthoryear{Bahdanau \bgroup \em et al.\egroup
  }{2014}]{bahdanau2014neural}
Dzmitry Bahdanau, Kyunghyun Cho, and Yoshua Bengio.
\newblock Neural machine translation by jointly learning to align and
  translate.
\newblock {\em arXiv:1409.0473}, 2014.

\bibitem[\protect\citeauthoryear{Bengio \bgroup \em et al.\egroup
  }{1994}]{bengio1994learning}
Yoshua Bengio, Patrice Simard, and Paolo Frasconi.
\newblock Learning long-term dependencies with gradient descent is difficult.
\newblock {\em IEEE Transactions on Neural Networks}, 5(2):157--166, 1994.

\bibitem[\protect\citeauthoryear{Bouchachia and
  Bouchachia}{2008}]{Bouchachia2008}
Abdelhamid Bouchachia and Saliha Bouchachia.
\newblock Ensemble learning for time series prediction.
\newblock In {\em Proceedings of the 1st International Workshop on Nonlinear
  Dynamics and Synchronization}, 2008.

\bibitem[\protect\citeauthoryear{Brockwell and Davis}{2009}]{Brockwell2009}
Peter~J. Brockwell and Richard~A Davis.
\newblock {\em Time Series: Theory and Methods (2nd ed.)}.
\newblock Springer, 2009.

\bibitem[\protect\citeauthoryear{Chakraborty \bgroup \em et al.\egroup
  }{2012}]{chakraborty2012fine}
Prithwish Chakraborty, Manish Marwah, Martin~F Arlitt, and Naren Ramakrishnan.
\newblock Fine-grained photovoltaic output prediction using a bayesian
  ensemble.
\newblock In {\em AAAI}, 2012.

\bibitem[\protect\citeauthoryear{Chen \bgroup \em et al.\egroup
  }{2008}]{Chen2008}
S.~Chen, X.~X. Wang, and C.~J. Harris.
\newblock Narx-based nonlinear system identification using orthogonal least
  squares basis hunting.
\newblock {\em IEEE Transactions on Control Systems Technology}, 16(1):78--84,
  2008.

\bibitem[\protect\citeauthoryear{Cho \bgroup \em et al.\egroup
  }{2014a}]{cho2014properties}
Kyunghyun Cho, Bart Van~Merri{\"e}nboer, Dzmitry Bahdanau, and Yoshua Bengio.
\newblock On the properties of neural machine translation: Encoder-decoder
  approaches.
\newblock {\em arXiv:1409.1259}, 2014.

\bibitem[\protect\citeauthoryear{Cho \bgroup \em et al.\egroup
  }{2014b}]{cho2014learning}
Kyunghyun Cho, Bart Van~Merri{\"e}nboer, Caglar Gulcehre, Dzmitry Bahdanau,
  Fethi Bougares, Holger Schwenk, and Yoshua Bengio.
\newblock Learning phrase representations using rnn encoder-decoder for
  statistical machine translation.
\newblock {\em arXiv:1406.1078}, 2014.

\bibitem[\protect\citeauthoryear{Diaconescu}{2008}]{Diaconescu08}
Eugen Diaconescu.
\newblock The use of {NARX} neural networks to predict chaotic time series.
\newblock {\em WSEA Transactions on Computer Research}, 3(3), 2008.

\bibitem[\protect\citeauthoryear{Elman}{1991}]{elman1991distributed}
Jeffrey~L Elman.
\newblock Distributed representations, simple recurrent networks, and
  grammatical structure.
\newblock {\em Machine learning}, 7(2-3):195--225, 1991.

\bibitem[\protect\citeauthoryear{Frigola and Rasmussen}{2014}]{Frigola2013}
R.~Frigola and C.~E. Rasmussen.
\newblock Integrated pre-processing for bayesian nonlinear system
  identification with gaussian processes.
\newblock In {\em IEEE Conference on Decision and Control}, pages 552--560,
  2014.

\bibitem[\protect\citeauthoryear{Gao and Er}{2005}]{gao2005narmax}
Yang Gao and Meng~Joo Er.
\newblock Narmax time series model prediction: feedforward and recurrent fuzzy
  neural network approaches.
\newblock {\em Fuzzy Sets and Systems}, 150(2):331--350, 2005.

\bibitem[\protect\citeauthoryear{Graves \bgroup \em et al.\egroup
  }{2013}]{graves2013speech}
Alex Graves, Abdel-rahman Mohamed, and Geoffrey Hinton.
\newblock Speech recognition with deep recurrent neural networks.
\newblock In {\em ICASSP}, pages 6645--6649, 2013.

\bibitem[\protect\citeauthoryear{Hochreiter and
  Schmidhuber}{1997}]{hochreiter1997long}
Sepp Hochreiter and J{\"u}rgen Schmidhuber.
\newblock Long short-term memory.
\newblock {\em Neural Computation}, 9(8):1735--1780, 1997.

\bibitem[\protect\citeauthoryear{H{\"u}bner \bgroup \em et al.\egroup
  }{2010}]{hubner2010dual}
Ronald H{\"u}bner, Marco Steinhauser, and Carola Lehle.
\newblock A dual-stage two-phase model of selective attention.
\newblock {\em Psychological Review}, 117(3):759--784, 2010.

\bibitem[\protect\citeauthoryear{Kalchbrenner and
  Blunsom}{2013}]{kalchbrenner2013recurrent}
Nal Kalchbrenner and Phil Blunsom.
\newblock Recurrent continuous translation models.
\newblock In {\em EMNLP}, volume~3, pages 413--422, 2013.

\bibitem[\protect\citeauthoryear{Karpathy and Li}{2015}]{karpathy2015deep}
Andrej Karpathy and Fei-Fei Li.
\newblock Deep visual-semantic alignments for generating image descriptions.
\newblock In {\em CVPR}, pages 3128--3137, 2015.

\bibitem[\protect\citeauthoryear{Kingma and Ba}{2014}]{kingma2014adam}
Diederik Kingma and Jimmy Ba.
\newblock Adam: A method for stochastic optimization.
\newblock {\em arXiv:1412.6980}, 2014.

\bibitem[\protect\citeauthoryear{Lin \bgroup \em et al.\egroup
  }{1996}]{lin1996learning}
Tsungnan Lin, Bill~G. Horne, Peter Tino, and C.~Lee Giles.
\newblock Learning long-term dependencies in {NARX} recurrent neural networks.
\newblock {\em IEEE Transactions on Neural Networks}, 7(6):1329--1338, 1996.

\bibitem[\protect\citeauthoryear{Liu and Hauskrecht}{2015}]{liu2015regularized}
Zitao Liu and Milos Hauskrecht.
\newblock A regularized linear dynamical system framework for multivariate time
  series analysis.
\newblock In {\em AAAI}, pages 1798--1805, 2015.

\bibitem[\protect\citeauthoryear{Plutowski \bgroup \em et al.\egroup
  }{1996}]{plutowski1996experience}
Mark Plutowski, Garrison Cottrell, and Halbert White.
\newblock Experience with selecting exemplars from clean data.
\newblock {\em Neural Networks}, 9(2):273--294, 1996.

\bibitem[\protect\citeauthoryear{Pu \bgroup \em et al.\egroup
  }{2016}]{pu2016adaptive}
Yunchen Pu, Martin~Renqiang Min, Zhe Gan, and Lawrence Carin.
\newblock Adaptive feature abstraction for translating video to language.
\newblock {\em arXiv preprint arXiv:1611.07837}, 2016.

\bibitem[\protect\citeauthoryear{Qin \bgroup \em et al.\egroup
  }{2015}]{qin2015saliency}
Yao Qin, Huchuan Lu, Yiqun Xu, and He~Wang.
\newblock Saliency detection via cellular automata.
\newblock In {\em CVPR}, pages 110--119, 2015.

\bibitem[\protect\citeauthoryear{Rumelhart \bgroup \em et al.\egroup
  }{1986}]{rumelhart1986learning}
David~E Rumelhart, Geoffrey~E Hinton, and Ronald~J Williams.
\newblock Learning representations by back-propagating errors.
\newblock {\em Nature}, 323(9):533--536, 1986.

\bibitem[\protect\citeauthoryear{Song \bgroup \em et al.\egroup
  }{2015}]{song2015top}
Dongjin Song, Wei Liu, Rongrong Ji, David~A Meyer, and John~R Smith.
\newblock Top rank supervised binary coding for visual search.
\newblock In {\em ICCV}, pages 1922--1930, 2015.

\bibitem[\protect\citeauthoryear{Song \bgroup \em et al.\egroup
  }{2016}]{song2016fast}
Dongjin Song, Wei Liu, and David~A Meyer.
\newblock Fast structural binary coding.
\newblock In {\em IJCAI}, pages 2018--2024, 2016.

\bibitem[\protect\citeauthoryear{Sutskever \bgroup \em et al.\egroup
  }{2014}]{sutskever2014sequence}
Ilya Sutskever, Oriol Vinyals, and Quoc~V Le.
\newblock Sequence to sequence learning with neural networks.
\newblock In {\em NIPS}, pages 3104--3112, 2014.

\bibitem[\protect\citeauthoryear{Werbos}{1990}]{werbos1990backpropagation}
Paul~J Werbos.
\newblock Backpropagation through time: what it does and how to do it.
\newblock {\em Proceedings of the IEEE}, 78(10):1550--1560, 1990.

\bibitem[\protect\citeauthoryear{Whittle}{1951}]{Whittle1951}
P.~Whittle.
\newblock {\em Hypothesis Testing in Time Series Analysis}.
\newblock PhD thesis, 1951.

\bibitem[\protect\citeauthoryear{Wu \bgroup \em et al.\egroup
  }{2013}]{wu2013dynamic}
Yue Wu, Jos{\'e}~Miguel Hern{\'a}ndez-Lobato, and Zoubin Ghahramani.
\newblock Dynamic covariance models for multivariate financial time series.
\newblock In {\em ICML}, pages 558--566, 2013.

\bibitem[\protect\citeauthoryear{Xu \bgroup \em et al.\egroup
  }{2015}]{xu2015show}
Kelvin Xu, Jimmy Ba, Ryan Kiros, Kyunghyun Cho, Aaron~C Courville, Ruslan
  Salakhutdinov, Richard~S Zemel, and Yoshua Bengio.
\newblock Show, attend and tell: Neural image caption generation with visual
  attention.
\newblock In {\em ICML}, volume~14, pages 77--81, 2015.

\bibitem[\protect\citeauthoryear{Yan \bgroup \em et al.\egroup
  }{2013}]{Cottrell-NARX}
Linjun Yan, Ahmed Elgamal, and Garrison~W. Cottrell.
\newblock Substructure vibration {NARX} neural network approach for statistical
  damage inference.
\newblock {\em Journal of Engineering Mechanics}, 139:737--747, 2013.

\bibitem[\protect\citeauthoryear{Yang \bgroup \em et al.\egroup
  }{2016}]{Yang2016}
Zichao Yang, Diyi Yang, Chris Dyer, Xiaodong He, Alex Smola, and Eduard Hovy.
\newblock Hierarchical attention networks for document classification.
\newblock In {\em NAACL}, 2016.

\end{thebibliography}

\end{document}